\documentclass{article}
\usepackage{PRIMEarxiv}
\usepackage{graphicx}
\usepackage{subcaption}
\usepackage{float}
\usepackage{algpseudocode}
\usepackage{amsmath}
\usepackage{amsfonts}
\usepackage{amssymb}
\usepackage{multirow}
\usepackage{multicol}
\usepackage{graphicx}
\usepackage{caption}
\usepackage{subcaption}
\usepackage[T1]{fontenc}
\usepackage{comment}
\usepackage{booktabs}
\usepackage{pifont}
\usepackage[linesnumbered,ruled,vlined]{algorithm2e}
\usepackage{wrapfig}
\usepackage{xcolor}
\usepackage{graphicx}
\usepackage{xspace}
\usepackage{eso-pic}
\usepackage[T1]{fontenc}
\usepackage{mathptmx}
\usepackage{helvet,courier}
\usepackage{mathtools}
\usepackage{bm}

\title{Vanishing Activations: A Symptom of Deep Capsule Networks}

\author{Miles Everett, Mingjun Zhong and Georgios Leontidis \\
 Department of Computing Science \\
University of Aberdeen \\
Aberdeen, United Kingdom \\
\texttt{\{m.everett.20, mingjun.zhong, georgios.leontidis\}@abdn.ac.uk} \\
}

\begin{document}

\maketitle

\begin{abstract}

Capsule Networks, an extension to Neural Networks utilizing vector or matrix representations instead of scalars, were initially developed to create a dynamic parse tree where visual concepts evolve from parts to complete objects. Early implementations of Capsule Networks achieved and maintain state-of-the-art results on various datasets. However, recent studies have revealed shortcomings in the original Capsule Network architecture, notably its failure to construct a parse tree and its susceptibility to vanishing gradients when deployed in deeper networks. This paper extends the investigation to a range of leading Capsule Network architectures, demonstrating that these issues are not confined to the original design. We argue that the majority of Capsule Network research has produced architectures that, while modestly divergent from the original Capsule Network, still retain a fundamentally similar structure. We posit that this inherent design similarity might be impeding the scalability of Capsule Networks. Our study contributes to the broader discussion on improving the robustness and scalability of Capsule Networks.

\end{abstract}

%------------------------------------------------------------------------- 
\section{Introduction}
\label{sec:intro}

% - Since Capsule Networks inception, the majority of research has been focused on improving the routing algorithm and testing on toy benchmarks.

% - Very little research has been put into scaling Capsule Networks to more challenging datasets.

% - Recent paper: "Why Capsule Neural Networks Do Not Scale: Challenging the Dynamic Parse-Tree Assumption" showed that Dynamically Routed Capsule Networks with multiple layers suffer from low network utitilisation.

% - We deemed it necessary to investigate whether this is an issue that only effects Dynamic Routing Capsule Networks or extends across the entire field.

Since their inception, Capsule Networks \cite{sabour2017dynamic} have drawn significant attention in the field of Neural Networks. By utilizing vector or matrix \cite{hinton2018matrix} representations instead of scalars, these networks propose the idea of creating a dynamic parse tree where visual concepts transition from parts to full objects as they pass through Capsule layers. Yet, despite the initial enthusiasm, the interest in Capsule Networks has waned over time, primarily due to challenges associated with scaling.

The majority of research efforts to date have focused on refining the routing algorithm of Capsule Networks \cite{hinton2018matrix, ribeiro2020capsule, hahn2019self} and evaluating their performance on simple, toy benchmarks. However, in-depth exploration of these new routing algorthims ability to advance Capsule Networks' scalability, particularly with more complex and challenging datasets, has been conspicuously lacking. This deficit in understanding leaves a gap in our knowledge about the networks' full potential and their applicability in broader, real-world contexts.

A recent study titled "Why Capsule Neural Networks Do Not Scale: Challenging the Dynamic Parse-Tree Assumption" \cite{mitterreiter2023capsule} has shed light on a critical issue related to this problem. The authors found that Dynamically Routed Capsule Networks suffer from low network utilization as a result of a vanishing gradient, significantly limiting their scalability, especially as the problem gets more significant when more layers are added. This finding has potential to have big implications for the continued development and use of Capsule Networks.

Motivated by these findings, our study extends the investigation to other Capsule Network architectures. We seek to determine whether the issue of low network utilization is confined to Dynamically Routed Capsule Networks or if it is a more pervasive problem that affects the entire spectrum of Capsule Network architectures.

\subsection{Motivation and Contributions}

While Capsule Networks have been shown not to scale, they still maintain state of the art performance on toy datasets which test a networks ability to generalise to unseen viewpoints and instances of a class at test time. So while Capsule Networks in their current form aren't able to handle modern datasets, if they can be modified they could be extremely competitive with state of the art architectures like Vision Transformers \cite{dosovitskiy2020image, khan2022transformers} which are known to require a large amount of data for best performance.

We put forward the hypothesis that the stagnation and eventual degradation in test accuracy, especially when the number of layers becomes higher, is potentially due to the lack of Capsule activation. Capsule Networks are designed to recognize and preserve hierarchical relationships in data; however, our hypothesis suggests that the failure of Capsules to adequately activate may prevent the network from fully exploiting these hierarchies, thereby curtailing the expected performance increase.

Our research aims to provide an improved understanding of the current limitations of current Capsule Network architectures. We believe this understanding is a crucial step toward creating a future path for research in Capsule Networks that could eventually overcome these scalability issues.

\section{Background and Related Works}

\subsection{Concept of a Capsule}
% A capsule is a group of neurons that work together to identify the presence of a specific feature or entity within the input data. Unlike individual neurons in a CNN, which are scalar and detect the presence of specific features independently, capsules are vector or matrix representations that aim to capture the instantiation parameters (e.g., pose, scale, orientation) of an entity in a coordinated manner. This enables the network to better represent hierarchical relationships and maintain viewpoint invariance.

A capsule is a cluster of neurons in a neural network \cite{ribeiro2022learning}. Unlike scalar neurons in CNNs, capsules are vector or matrix representations that identify specific features in input data, capturing instantiation parameters like pose, scale, or orientation.

% Routing is the process by which capsule networks determine the hierarchical relationships between capsules in different layers of the network. In Dynamically Routed Capsule Networks for example, the network learns to route the output of lower-level capsules to appropriate higher-level capsules through a soft assignment mechanism. This allows the network to build a more coherent representation of the input data by encouraging high-level capsules to be activated only by lower-level capsules that agree on the presence of a specific entity.

Rather than using pooling operations found in CNNs, Capsule Networks introduce the concept of routing, which is the method capsule networks use to map relationships between capsules across different layers. It uses a soft assignment mechanism, allowing the network to create a consistent representation of the input data by calculating agreement between lower and higher-level capsules \cite{de2020introducing}.

% Capsule activation refers to the degree to which a capsule is activated or "engaged" in response to the input data. A higher value indicates a higher activation level, suggesting that the capsule has detected the presence of its corresponding entity with high confidence. In contrast, a lower value indicates low activation and, thus, lower confidence.

Capsule activation is the measure of a capsule's engagement between 0 and 1 in response to input data. A high value denotes high activation, indicating the capsule has confidently detected its corresponding entity, while a low value signals low confidence. Unlike routing, the sum of Capsule activations across a layer do not need to sum to 1.

\subsection{Dynamic Routing}

Dynamic Routing in Capsule Networks iteratively refines inter-capsule connections, introducing "coupling coefficients" ($c_{ij}$) to denote connection strength between capsules $i$ and $j$. The coefficients are updated based on the agreement ($a_{ij}$), computed as a dot product between a lower-level capsule ($u_i$) and a predicted output ($\hat{y}_{ij}$):

\begin{equation}
a_{ij} = \langle u_i, \hat{y}_{ij} \rangle = \langle u_i, W_{ij}u_i \rangle
\end{equation}

Coefficients are updated using a softmax function over the agreement values:

\begin{equation}
c_{ij} = \frac{\exp(a_{ij})}{\sum_k \exp(a_{ik})}
\end{equation}

After a fixed number of iterations, the output ($v_j$) of each higher-level capsule is the weighted sum of lower-level capsules' outputs:

\begin{equation}
v_j = \sum_i c_{ij} * \hat{y}_{ij}
\end{equation}

\subsection{EM Routing}

Expectation-Maximization (EM) Routing \cite{hinton2018matrix} is a routing mechanism which aims to route Capsules from a lower layer to a higher layer by leveraging the Expectation-Maximization algorithm, the EM Routing method iteratively refines routing assignments and updates high-level capsule activations, ultimately maximizing the likelihood of the input data given the capsule activations and routing assignments.

In the E-step, the posterior probabilities ($R_{ij}$) of the high-level capsules ($v_j$) being activated by the lower-level capsules ($u_i$) are calculated. These probabilities are updated according to the current capsule activations and the observed data:

\begin{equation}
R_{ij} = \frac{\exp(-\|u_i - W_{ij}v_j\|^2)}{\sum_k \exp(-\|u_i - W_{ik}v_k\|^2)}
\end{equation}

In the M-step, the high-level capsule activations ($v_j$) are updated based on the current routing assignments ($R_{ij}$) and lower-level capsules ($u_i$). The updated activations aim to minimize the weighted sum of squared differences between the lower-level capsules and their predictions:

\begin{equation}
v_j = \frac{\sum_i R_{ij}u_i}{\sum_i R_{ij}}
\end{equation}

The EM Routing algorithm alternates between the E-step and M-step until convergence or for a fixed number of iterations. By maximizing the likelihood of the input data and iteratively refining the routing assignments, EM Routing enhances the accuracy and robustness of Capsule Networks, particularly in tasks that involve complex, cluttered environments.

\subsection{Variational Bayes Routing}

Variational Bayes (VB) Routing \cite{ribeiro2020capsule} is a routing method introduced to overcome some of the challenges associated with EM Routing, including training instability and reproducibility. VB Routing applies Bayesian learning principles \cite{bishop2006pattern} to Capsule Networks by placing priors and modelling uncertainty over capsule parameters between layers.

The initial step in VB Routing assigns equal routing coefficients $\gamma_{ij}$ for all capsules in layers $\mathcal{L}_i$ and $\mathcal{L}_j$:
\begin{equation}
\forall \ i, j \ \text{capsules in layer} \ \mathcal{L}_i \ \text{and} \ \mathcal{L}_j: \gamma_{ij} \gets N_j^{-1}
\end{equation}

The next step, termed as 'voting', involves the computation of votes $\mathbf{V}_{j|i}$ for each capsule $i$ in $\mathcal{L}_i$ to capsule $j$ in $\mathcal{L}_j$ using a transformation matrix $\mathbf{W}_{ij}$ and the pose matrix $\mathbf{M}_{i}$:

\begin{equation}
\forall \ i, j \ : \mathbf{V}_{j|i} \gets \mathbf{M}_{i} \cdot \mathbf{W}_{ij}
\end{equation}

VB Routing then iterates $r$ times over two key steps: updating the routing weights and the $q^\star(\boldsymbol{\pi}, \boldsymbol{\mu}, \boldsymbol{\Lambda})$ parameters:

\begin{equation}
\forall \ i \in \mathcal{L}_i \ : \gamma_{ij} \leftarrow \gamma_{ij} \odot \bm{a}_i,
\end{equation}

\begin{equation}
\forall \ j \in \mathcal{L}_j \ : \ \text{Update} \ q^\star(\boldsymbol{\pi}, \boldsymbol{\mu}, \boldsymbol{\Lambda}),
\end{equation}

\begin{equation}
\forall \ i \in \mathcal{L}_i \ : \ \text{Update} \ q^\star(\bm{z}).
\end{equation}

At the end of these $r$ iterations, VB Routing updates the activation of each capsule $j$ in $\mathcal{L}_j$ using a logistic function $\sigma$:

\begin{equation}
\forall \ j \in \mathcal{L}_j \ : a'_j = \sigma(\beta_a - \big(\beta_u + \mathbb{E}[\ln \boldsymbol{\pi}_j] + \mathbb{E}[\ln \mathrm{det}(\boldsymbol{\Lambda}_j)])),
\end{equation}

where $\beta_a$ and $\beta_u$ are learnable parameters, $\boldsymbol{\pi}_j$ are the mixing coefficients, and $\boldsymbol{\Lambda}_j$ is the precision matrix.

\subsection{Self Routing}

% Self-Routing Capsule Networks (SR-CapsNet) \cite{hahn2019self} is a different approach that attempts to tackle the huge computational expense associated with routing. Inspired by the Mixture-of-Experts model, the authors of the Self Routing Capsule Networks proposed that each capsule independently determines its routing coefficients, with no need for agreement coordination with other capsules. Instead, each capsule uses a routing network designed to predict the routing coefficients directly.

% They propose that capsules often specialize in disjoint regions of the feature space, leading them to make multiple predictions based on the region-specific information available. At the layer level, this implies the existence of a collection of submodules that activate differently per example, akin to a mixture of experts where each expert specializes in different input space regions.

Self-Routing Capsule Networks (SR-CapsNet) \cite{hahn2019self} address the computational cost of routing by allowing each capsule to independently determine its routing coefficients, obviating the need for agreement coordination. Each capsule uses a routing network, inspired by mixture of experts \cite{masoudnia2014mixture} to predict routing coefficients directly. The model suggests that capsules specialize in disjoint regions of the feature space and make predictions based on region-specific information, functioning similarly to a mixture of experts with specializations in different input space regions.

In SR-CapsNet, the computation of the routing coefficients ($c_{ij}$) and predictions ($\widehat{\mathbf{u}}_{j|i}$) involves two learnable weight matrices, equivalent to a fully connected layer for each Capsule in the higher layer, $\mathbf{W}^{route}$ and $\mathbf{W}^{pose}$ respectively. For each layer of the routing network, each pose vector $\mathbf{u}_i$ is multiplied by a trainable weight matrix $\mathbf{W}^{route}$, which produces the routing coefficients directly. After softmax normalization, these coefficients are multiplied by the capsule's activation scalar $a_i$ to create weighted votes. The activation $a_j$ of a higher-layer capsule is the summation of these weighted votes from lower-level capsules over spatial dimensions $H \times W$, or $K \times K$ when using convolutions.

SR-CapsNet has exhibited competitive performance on standard capsule network benchmarks. However, this approach somewhat restricts the capsule network's ability to dynamically adjust routing weights based on the input, as these weights are now entirely dependent on the learned parameters of the routing subnetworks.

\section{Methodology}

To adequately evaluate the effect that number of layers has on activations within a Capsule Network, we need to evaluate the full range of depth that these architectures perform well at, but also to go beyond this depth to push the networks to breaking to allow for analysis.

\subsection{Architecture Choice}

We have chosen three different Capsule architectures from dozens of choices. EM Routing \cite{hinton2018matrix} was chosen as it is the first to extend the work of the original CapsNet \cite{sabour2017dynamic} and treats the activations as a separate scalar from the pose matrix for each Capsule. VB Routing \cite{ribeiro2020capsule} was chosen as the authors not only achieve SOTA results for Capsule Networks on some datasets, but they report a 4 layer Capsule Network as being their best performing, a deviation from the normal 0 or 1 layer Capsule Networks. Finally, Self Routing Capsule Networks \cite{hahn2019self} were chosen as their routing is entirely trained by backpropagation rather than clustering based upon the batch with only the transformation matrices being updated between steps.

For each of these architectures, we train networks consisting of 1-10 Convolutional Capsule layers. In our experiments, when we state that there is n number of Capsule layers, we do not include the Primary Capsules or Class Capsules in this count, so you may consider the number of layers to be the number of additional Capsule layers. E.g. a 1 layer Capsule Network connects the Primary Capsules to Convolutional Capsules and then to the Class Capsules, resulting in two instances of routing. 

We use a kernel size of 3, stride of 1 and padding of 1 and 16 Capsules whenever we have more than 1 Capsule layer to ensure that the feature map remains a consistent size. We fix the convolutional backbone to whatever it was defined as in the original implementations to stay true to the original implementations as much as possible. For implementation of networks with many layers, we employ activation checkpointing \cite{chen2016training} to avoid out of memory errors. We train for 50 epochs as this was determined to be sufficient for Capsule Networks to no longer make any improvements by training further.

\subsection{Datasets}

We choose 5 datasets for our experimentation which are MNIST, FashionMNIST, Cifar10, SmallNORB and ImageWoof. The first three are standard image classification benchmarks of increasing difficulty, but remain within the difficulty levels that Capsule Networks excel at. SmallNORB is not a difficult dataset to train on, but presents a unique problem in that the test set has been specifically chosen to test a networks ability to generalise to unseen viewpoints and new instances of the same class. Finally, we choose ImageWoof. This dataset contains 10 classes of different breeds of dogs from the ImageNet dataset. These images have high resolution and the added challenge of all being subclasses of the dog superclass, and thus having similar elements. Besides normalising the images using the training set means and standard deviations, we do not use any other kind of augmentations.

\subsection{Defining Dead Capsules}

% - During training, we track activations of each Capsule layer during validation and testing epochs.
% - We average the activations over the feature map and batch dimensions to obtain an average activation for each Capsule over the entire epoch.
% - We consider any Capsule with an average activation of less than 0.01 to be dead and not contributing to the network.

% - How average activation will be calculated
% - Defining concept of dead capsules

In order to gain insight into the behavior of the Capsule Network during the training process, we monitored the activations of each Capsule layer during the validation and testing epochs. Specifically, we denote the activation of the $i$-th Capsule in the $j$-th batch and the $k$-th feature map as $a_{jik}$, the process can be formally written as:

\begin{equation}
A_{i} = \frac{1}{N \times M} \sum_{j=1}^{N} \sum_{k=1}^{M} a_{jik},
\end{equation}
where $N$ is the number of batches and $M$ is the number of feature maps. $A_{i}$ represents the average activation of the $i$-th Capsule over the entire epoch. Note that we combine the height and width dimensions of the feature map into M for ease of notation.

Upon obtaining the average activations, we then inspected each Capsule's contribution to the network. A Capsule was deemed "dead" or non-contributing if its average activation was less than or equal to a threshold of 0.01, 

\begin{equation}
A_{i} <= 0.01.
\end{equation}
This decision was based on the assumption that a Capsule with such a low average activation was not significantly contributing to the final output of the network. In subsequent analyses and discussions, the average number of these "dead" Capsules are examined.

\section{Capsule Networks Do Not Scale}

% - This section should be purely results of the experimentation

% In this study, we demonstrate that when additional layers are incorporated into a Capsule Network, the typical performance improvements observed in other neural network architectures are not realized. Unlike in conventional networks where additional layers often contribute to enhanced depth and complexity, thereby improving performance, Capsule Networks do not demonstrate a similar increase.

% Our findings also corroborate and extend the phenomenon reported in \cite{mitterreiter2023capsule}. We show that the scalability issues persist beyond Dynamic Routing, reaching into other routing algorithms including EM Routing, Self Routing, and VB Routing. This indicates that the problem is more deeply ingrained in the architecture rather than being a consequence of a particular routing algorithm.

In this section, we will show the results of our experimentation, highlighting the alarming amount of "dead" Capsules, referring to those Capsules with less than or equal to 0.01\% activation on average for the entire test set. 

For each of our datasets, we show the test accuracy of each of our chosen Capsule architectures along with the average number of dead Capsules across the Convolutional Capsules layers. 

\begin{figure}[H]
    \centering
    \includegraphics[width=0.75\textwidth]{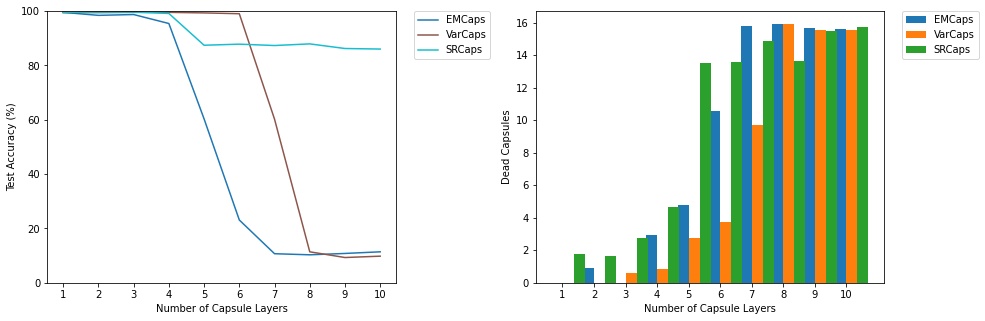}
    \vspace{5pt}
    \caption{Test Accuracy and average number of dead capsules across all layers for 1-10 ConvCapsule layers of EMCaps, VarCaps, and SRCaps for the MNIST Dataset \cite{lecun2010mnist}}
    \label{fig:side_by_side}
\end{figure}

\begin{figure}[H]
    \centering
    \includegraphics[width=0.75\textwidth]{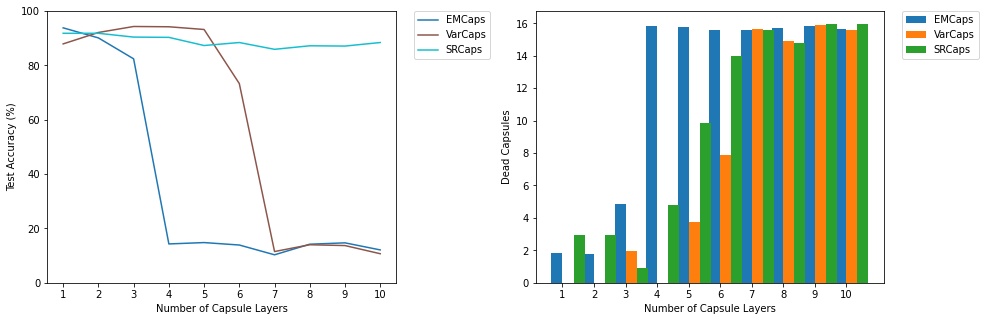}
    \vspace{5pt}
    \caption{Test Accuracy and average number of dead capsules across all layers for 1-10 ConvCapsule layers of EMCaps, VarCaps, and SRCaps for the Fashion-MNIST Dataset \cite{xiao2017fashion}}
    \label{fig:side_by_side}
\end{figure}

\begin{figure}[H]
    \centering
    \includegraphics[width=0.75\textwidth]{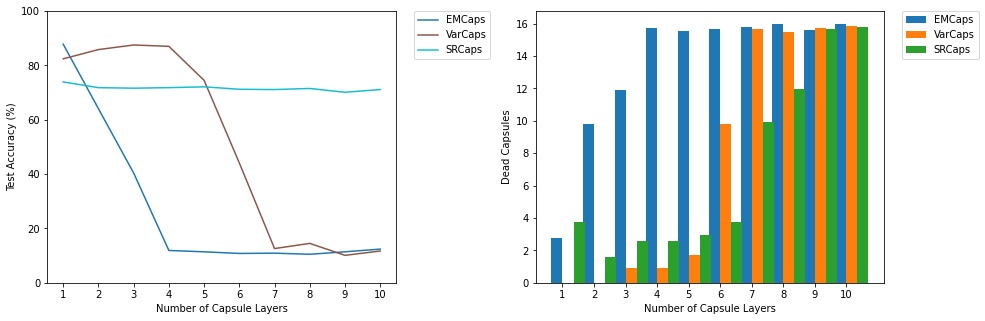}
    \vspace{5pt}
    \caption{Test Accuracy and average number of dead capsules across all layers for 1-10 ConvCapsule layers of EMCaps, VarCaps, and SRCaps for the Cifar-10 Dataset \cite{krizhevsky2009learning}}
    \label{fig:side_by_side}
\end{figure}

\begin{figure}[H]
    \centering
    \includegraphics[width=0.75\textwidth]{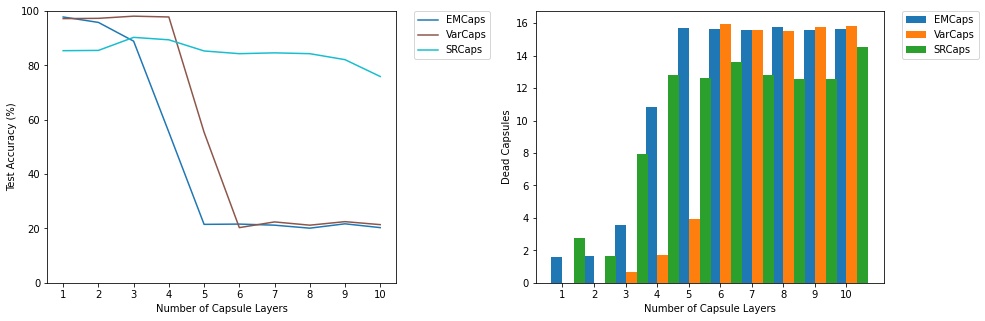}
    \vspace{5pt}
    \caption{Test Accuracy and average number of dead capsules across all layers for 1-10 ConvCapsule layers of EMCaps, VarCaps, and SRCaps for the SmallNORB Dataset \cite{lecun2004learning}}
    \label{fig:side_by_side}
\end{figure}

\begin{figure}[H]
    \centering
    \includegraphics[width=0.75\textwidth]{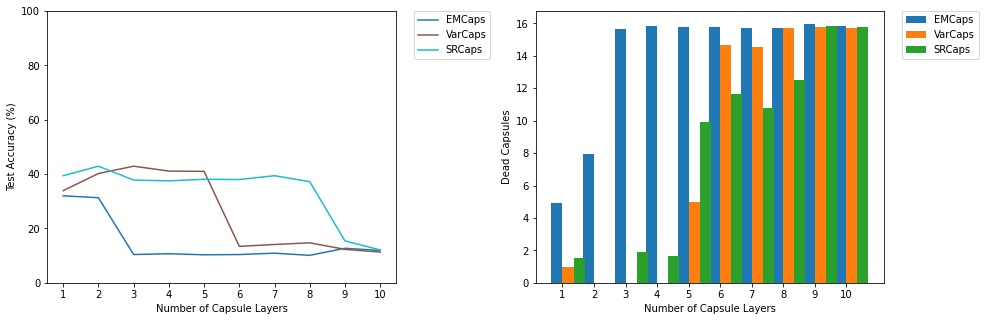}
    \vspace{5pt}
    \caption{Test Accuracy and average number of dead capsules across all layers for 1-10 ConvCapsule layers of EMCaps, VarCaps, and SRCaps for the ImageWoof Dataset \cite{Howard_Imagewoof_2019}}
    \label{fig:side_by_side}
\end{figure}

As we can see, there is a consistent pattern of more Capsules dying as we go deeper in our network, along with a clear relationship between the number of dead Capsules and the ability for the network to score a good test accuracy.

\subsection{Capsules Die During Training}

\begin{figure}[H]
    \centering
    \begin{subfigure}{0.49\textwidth}
        \includegraphics[width=\linewidth]{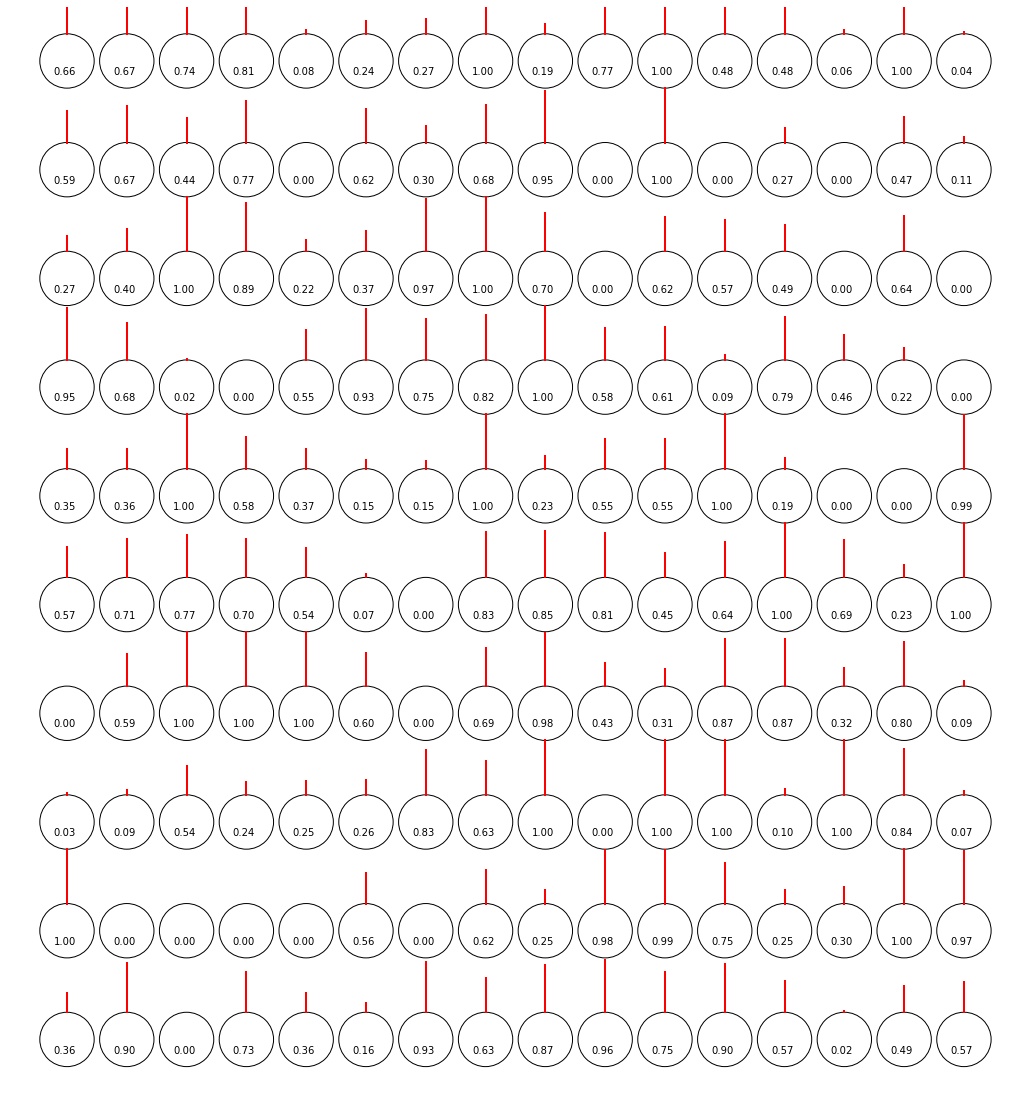}
        \caption{Individual Average Capsule Activations for 10 layer EM Routing \cite{hinton2018matrix} on the ImageWoof \cite{Howard_Imagewoof_2019} dataset at the end of epoch 1.}
    \end{subfigure}
    \hfill
    \begin{subfigure}{0.49\textwidth}
        \includegraphics[width=\linewidth]{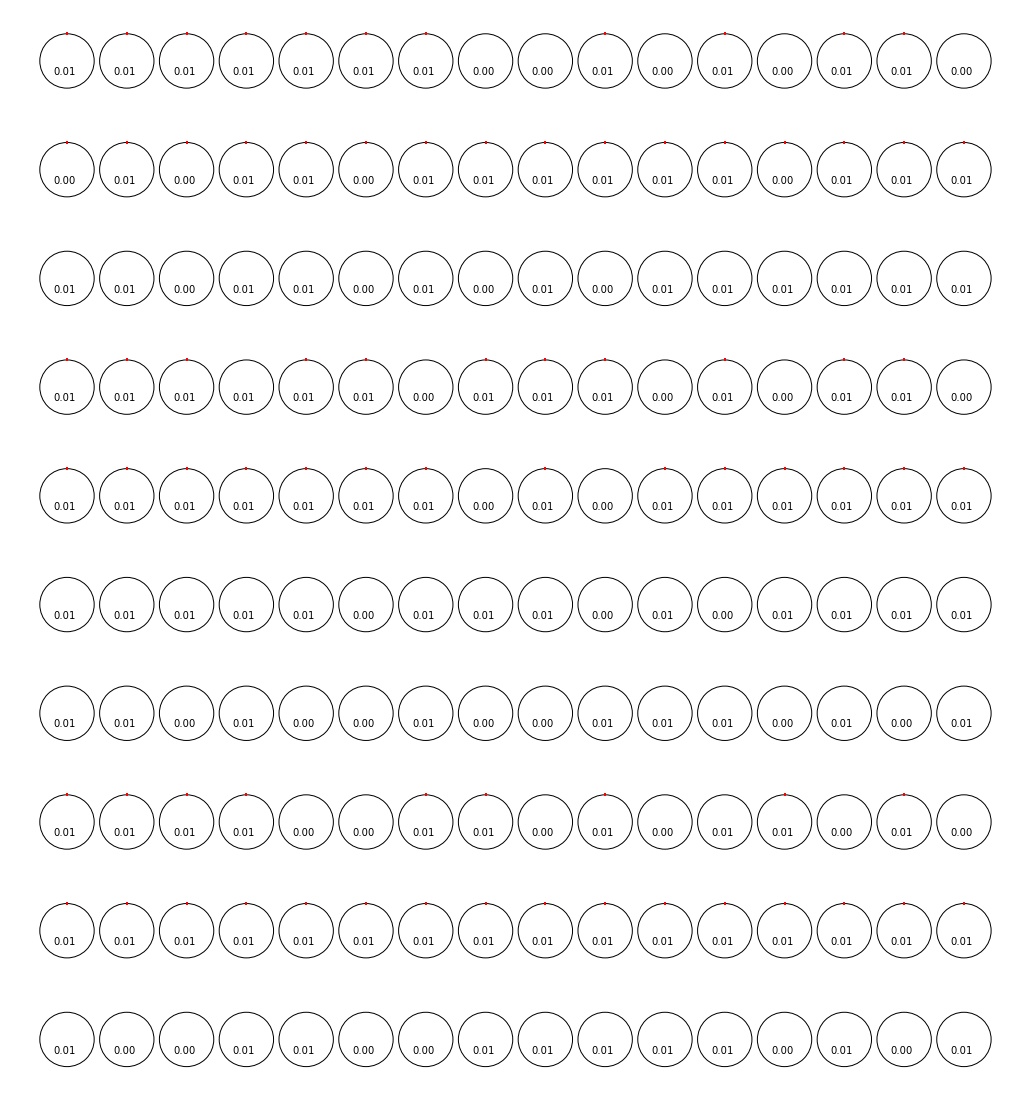}
        \caption{Individual Average Capsule Activations for 10 layer EM Routing \cite{hinton2018matrix} on the ImageWoof \cite{Howard_Imagewoof_2019} dataset at the end of epoch 50.}
    \end{subfigure}
    \vspace{5pt}
    \caption{Visualisation of how Capsule Activations are being killed as a consequence of training in a 10 layer Network. Each circle represents a Capsule with each row representing a layer of Capsules. These are the average activations across the entire validation epoch.}
\end{figure}

\begin{figure}[H]
    \centering
    \begin{subfigure}{0.49\textwidth}
        \includegraphics[width=\linewidth]{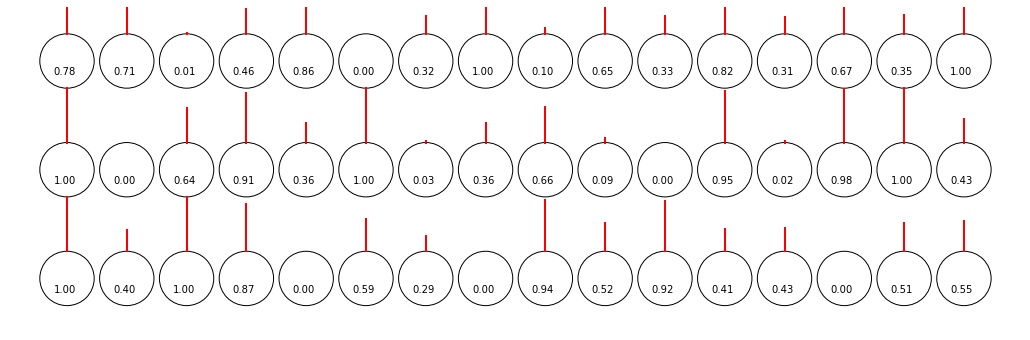}
        \caption{Individual Average Capsule Activations for 3 layer EM Routing \cite{hinton2018matrix} on the MNIST \cite{lecun2010mnist} dataset at the end of epoch 1.}
    \end{subfigure}
    \hfill
    \begin{subfigure}{0.49\textwidth}
        \includegraphics[width=\linewidth]{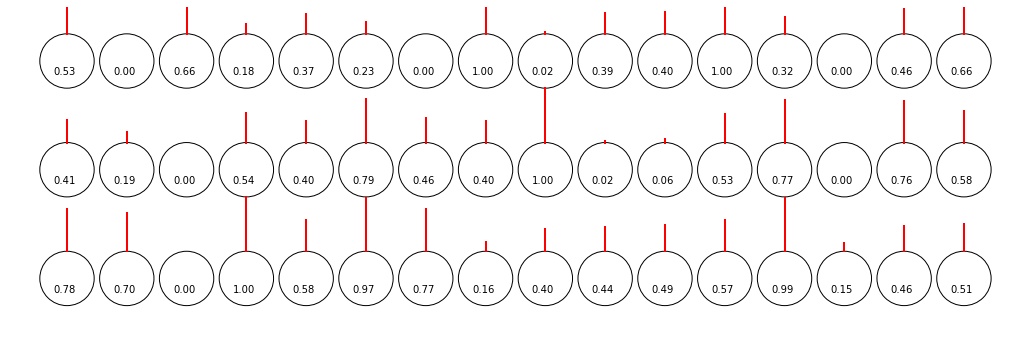}
        \caption{Individual Average Capsule Activations for 3 layer EM Routing \cite{hinton2018matrix} on the MNIST \cite{lecun2010mnist} dataset at the end of epoch 50.}
    \end{subfigure}
    \vspace{5pt}
    \caption{Visualisation of how Capsule Activations are changing throughout sequence of training in a 3 layer EM Network. Each circle represents a Capsule with each row representing a layer of Capsules. These are the average activations across the entire validation epoch.}
\end{figure}

Along with results on the test set of our datasets. We also track activations of our Capsules on validation epochs. In figure 6 we show how Capsules do not start off dead, but become dead throughout training. On the first validation epoch of training, we have a variety of "dead" Capsules and highly active Capsules, along with many Capsules activations fitting in somewhere inbetween. In stark contrast, by the time we reach epoch 50, all of our Capsules are completely dead, indicating a complete failure to learn, which is reflected in the 11.9\% test accuracy. Since Capsule activation values correspond to the confidence of the network that the concept that the Capsule represents are present, when there is no, or very little, activations from any Capsules it indicates that the Capsules are or are not able to learn the concepts from the information that they are fed. Figure 7 on the other hand shows behaviour that we would want to see, Capsule activation is not lost throughout training, and subsequently achieves 98.7\% accuracy.

\section{Conclusion}

% - Our analysis finds that the studied Capsule Networks have an optimal amount of layers, usually a low amount. 

% - Self Routing Capsule Networks appears to fare the best in terms of not degrading, but does not improve.

% - There is a point when training each architecture where the vanishing gradient problem presents itself and the network fails to learn anything due to lack of information flow meaning that the test accuracy becomes 1/n where n is the number of classes.

In conclusion, our analysis reveals some insightful patterns across the Capsule Network architectures we have studied. We found that these architectures typically have an optimal number of layers, which tends to be relatively low. This observation suggests a possible limitation in the maximum information that Capsule Networks can handle effectively in their current formulation, hence affecting their scalability and performance on more complex datasets.

Our findings also corroborate and extend the phenomenon reported in \cite{mitterreiter2023capsule}. We show that the scalability issues persist beyond Dynamic Routing, affecting other routing algorithms including EM Routing, Self Routing, and VB Routing. This indicates that the problem is more deeply ingrained in the architecture rather than being a consequence of a particular routing algorithm.

% In section 2, we showed how each of the routing algorithms controls the flow of information to the next layer of Capsules, with each algorithm involving a multiplication with the activation values. Thus we believe that as a result of the activation values often being very small, this limits the flow of information that can be passed between layers, and multiple stacked layers exacerbate this problem. In \cite{gugglberger2021training} they apply residual connections between stacked layers of Capsule Networks which allows them to train deeper networks with less of a dropoff compared to our results, but do not perform any analysis on the effect that this has on the internals of the Networks or apply the technique to a wider range of architectures.

As detailed in Section 2, each routing algorithm controls the information flow to the next layer of capsules by at some point utilising a multiplication of the pose matrices with the activation values. Given that these activation values are often small, they might limit information transfer between layers, especially in networks with multiple stacked layers. In \cite{gugglberger2021training}, the authors leverage residual skip connections between stacked layers of Capsule Networks, facilitating training of deeper networks with less performance drop-off relative to our results, although not an increase in performance. However, they do not thoroughly examine the effects of this method on the network's internal dynamics or apply it to a diverse range of architectures. Hence, we believe that further analysis utilizing this technique and others that have improved information flow in CNNs could be a fruitful avenue for future research.

Our study also showed that among the architectures under consideration, Self Routing Capsule Networks exhibited the most resilience. While they did not show improvements with the addition of layers, their performance did not degrade as noticeably as others. This relative stability might be indicative of some inherent robustness in the Self Routing mechanism, which is not reliant on a clustering mechanism, which could serve as a valuable insight for future research.

Additionally, a significant bottleneck we identified across all studied architectures is the emergence of the vanishing gradient problem. During the training phase, there exists a critical network size beyond which the network ceases to learn effectively. We attribute this to the lack of gradient flow, resulting in the network's inability to further optimize its parameters. As a consequence, the test accuracy for these networks drops to a baseline level of 1/n, where $n$ denotes the number of classes, indicating a failure to learn.

These findings show that simply stacking Capsule layers does not work, as was found in the early days of CNNs and MLPs. So we would consider the next steps to be evaluating whether techniques which were found to be effective for these networks can help Capsule Networks or whether a novel solution is needed to fix the flow of information issues.

%------------------------------------------------------------------------- 

%------------------------------------------------------------------------- 

% \subsection{References}

% List and number all bibliographical references in 9-point Times,
% single-spaced, at the end of your paper. When referenced in the text,
% enclose the citation number in square brackets, for
% example~\cite{Authors06}.  Where appropriate, include the name(s) of
% editors of referenced books.

\bibliographystyle{unsrt}

\end{document}